%% file: socia.tex
\documentclass[sigconf]{acmart}
\AtBeginDocument{%
  }

\usepackage{balance} 
\usepackage{xspace} 
\usepackage{graphicx}
\usepackage{amsmath}
\usepackage[most]{tcolorbox}
\usepackage{tabularx}
\usepackage{amsmath}      
\usepackage{algpseudocode} 
\usepackage{algorithm} 
\usepackage{multirow}
\usepackage{booktabs}    
\usepackage{colortbl}    
\usepackage{xcolor}    

\usepackage{tikz}
\usetikzlibrary{shapes,arrows,positioning,fit,backgrounds}
\usepackage{pgfplots}
\usepackage{pgfplotstable}
\usepgfplotslibrary{groupplots} 
\definecolor{softgray}{gray}{0.95}
\usetikzlibrary{plotmarks}
\usetikzlibrary{calc}

\newcommand{\socia}{\texttt{SOCIA}-\texttt{$\nabla$}\xspace}
\newcommand{\ofsocia}{\texttt{SOCIA}-\texttt{$\nabla$}’s\xspace}

\usepackage[dvipsnames,table]{xcolor}
\definecolor{OIOrange}{HTML}{D55E00} 
\definecolor{OIBlueGreen}{HTML}{009E73} 

\newcommand{\pos}[1]{\textcolor{black!75}{#1}}
\newcommand{\negv}[1]{\textcolor{teal!65!black}{#1}}

\newtcolorbox{simstepbox}{
  colback=orange!5!white,
  colframe=orange!60!black,
  coltitle=black,      
  boxrule=0.7pt,        
  arc=1pt,              
  outer arc=1pt,
  left=2pt, right=2pt, top=2pt, bottom=2pt,
  enhanced,             
  breakable
}

\copyrightyear{2026}
\acmYear{2026}
\setcopyright{cc}
\setcctype{by}
\acmConference[SIGSPATIAL '26]{The 34th ACM International Conference on Advances in Geographic Information Systems}{November 03--06, 2026}{Riverside, CA, USA}
\acmBooktitle{The 34th ACM International Conference on Advances in Geographic Information Systems (SIGSPATIAL '26), November 03--06, 2026, Riverside, CA, USA}
\acmDOI{10.1145/3841645.3844183}
\acmISBN{979-8-4007-2950-8/2026/11}

\begin{document}

\title[\socia: Automated Spatially Grounded Simulator Generation]
{\raisebox{-0.4ex}{\includegraphics[height=1.0em]{latex/figure/socia\_logo.png}}
\hspace{0.3em}\socia: Textual Gradient Meets Multi-Agent Orchestration for Automated Spatially Grounded Simulator Generation}

\author{Yuncheng Hua}
\affiliation{%
  \institution{University of New South Wales}
  \city{Sydney}
  \state{NSW}
  \country{Australia}
}
\email{devin.hua@unsw.edu.au}

\author{Sion Weatherhead}
\affiliation{%
  \institution{University of New South Wales}
  \city{Sydney}
  \state{NSW}
  \country{Australia}
}
\email{s.weatherhead@unsw.edu.au}

\author{Mehdi Jafari}
\affiliation{%
  \institution{University of New South Wales}
  \city{Sydney}
  \state{NSW}
  \country{Australia}
}
\email{mehdi.jafari@unsw.edu.au}

\author{Hao Xue}
\affiliation{%
  \institution{University of New South Wales}
  \city{Sydney}
  \state{NSW}
  \country{Australia}
}
\email{hao.xue1@unsw.edu.au}

\author{Flora D. Salim}
\affiliation{%
  \institution{University of New South Wales}
  \city{Sydney}
  \state{NSW}
  \country{Australia}
}
\email{flora.salim@unsw.edu.au}

\renewcommand{\shortauthors}{Hua et al.}

\begin{abstract}
\input{latex/chapter/0_abstract}
\end{abstract}





\settopmatter{authorsperrow=3}
\maketitle

\section{Introduction}
\label{sec:intro}
\input{latex/chapter/1_intro}

\section{Related Work}
\label{sec:related}
\input{latex/chapter/2_related}

\section{\socia: Approach}
\label{sec:approach}
\input{latex/chapter/3_approach}

\section{\socia: Experimental Setup}
\label{sec:exp_settings}
\input{latex/chapter/4_exp_setup}

\section{\socia: Evaluation}
\label{sec:evaluation}
\input{latex/chapter/5_exp_evaluation}

\section{Conclusion, Limitations, and Future Work}
\label{sec:conclusion}
\input{latex/chapter/6_conclusion}


\section*{AI Use Disclosure}
GPT-5.2 was used as the backbone language model for the agents in \socia and for the experiments reported in this paper.
ChatGPT was also used for language polishing and grammar correction during manuscript preparation.
No AI system was used to independently determine the scientific claims, interpret experimental results, or make final decisions regarding the technical content of the manuscript.
The authors reviewed, verified, and take full responsibility for all technical content, experimental results, analyses, citations, and final manuscript text.

\begin{acks}
\input{latex/chapter/acknow}
\end{acks}

\balance
\bibliographystyle{ACM-Reference-Format}
\bibliography{socia}

\clearpage
\appendix
\section{Appendix}
\subsection{Generated Simulator Structures}
\label{appen:simulator_examples}
\input{latex/supplementary/simulator_overview}

\subsection{Details of Ablation and Backbone Sensitivity}
\label{appen:ablation_details}
\input{latex/supplementary/ablation_details}

\subsection{Supplementary Experiments}
\label{appen:add_exps}
\input{latex/supplementary/0_supple}


\subsection{Representative Diagnosis-and-Repair Case Study}
\label{appen:diagnosis_case}

\input{latex/supplementary/2_case_study}

\end{document}

%% file: latex/chapter/0_abstract.tex

In this paper, we present \socia, a diagnosis-driven multi-agent framework for automated simulator-code construction. \socia iteratively refines executable simulator code within a textual computation graph, where LLM-driven agents synthesize, execute, evaluate, diagnose, and repair simulators. A grounded simulator blueprint attributes behavioral residuals to candidate mechanisms and their spatial or interaction contexts, which are then converted into localized textual repair signals with history-aware memory and constraint projection. We evaluate \socia on the real-world LLMob personal-mobility benchmark and a controlled semi-synthetic geo-social mask-adoption benchmark. Across in-distribution simulation, out-of-distribution what-if generation, and geo-social diffusion, \socia achieves the strongest overall simulator fidelity. \ofsocia code and data are publicly available \href{https://github.com/cruiseresearchgroup/SOCIA-Nabla}{here}.

%% file: latex/chapter/1_intro.tex
\begin{figure*}[t]
\centering
\includegraphics[width=1.0\textwidth]{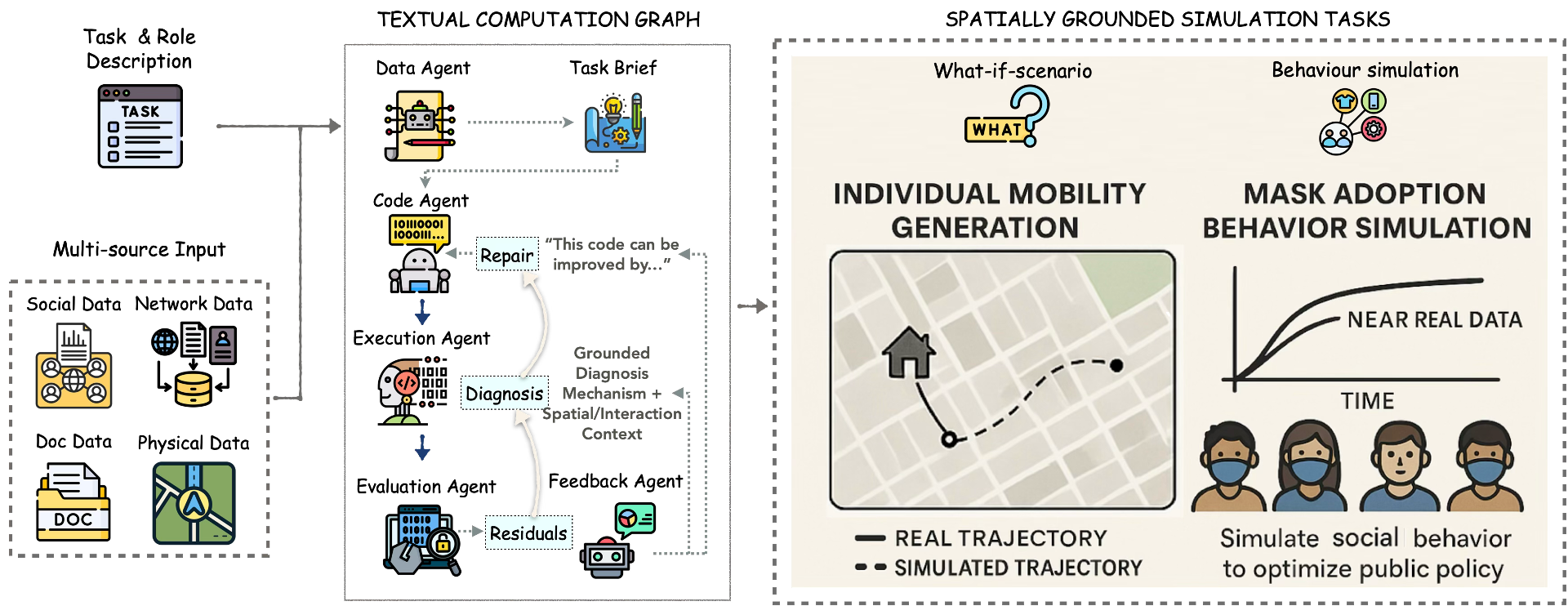} 
\caption{
Overview of \socia. 
The framework iteratively generates, executes, evaluates, diagnoses, and repairs simulator code. During backward repair, metric residuals are attributed to candidate simulator mechanisms and grounded in their relevant spatial or interaction contexts, producing localized repair signals. Right: personal mobility simulation and geo-social mask-adoption diffusion.
Third-party icons \copyright~Iconjam, Magnific, Uniconlabs, wanicon, zero\_wing, agus raharjo, Rusmicon, Catkuro, and Eucalyp, via Flaticon, used under the Flaticon Free License.
}
\label{fig:gradient}
\end{figure*}

Spatio-temporal and geo-social simulators are increasingly important for studying human mobility, public-health diffusion, urban interventions, and ``what-if'' decision support~\cite{DBLP:journals/corr/abs-2502-08691}. 
They provide low-cost, reproducible in-silico sandboxes for controlled intervention analysis, while enabling researchers to probe emergent outcomes under shifts and unseen regimes~\cite{bonabeau2002agent,cohen2017scientific,alves2023policy}. 
By design, simulators span two paradigms: \emph{policy-oriented aggregate (system-dynamics / compartmental) models}, which summarize populations through equations over aggregated state variables, and \emph{agent-based / microsimulation models}, which follow heterogeneous decision units and their interactions to capture fine-grained dynamics~\cite{van1998agent,mitton2000microsimulation,macal2005tutorial,howick2024framework}. 
These approaches arise across cyber, physical, and social (CPS) domains~\cite{DBLP:journals/epjds/RenTSCS18,DBLP:journals/tkde/RenTSCCS18,DBLP:journals/tosn/KaurSRCTS20,DBLP:journals/access/PasandidehPG22,mcculloch2022calibrating}, and are especially valuable in spatially grounded settings where mobility, interaction, information diffusion, and intervention effects unfold over time.

For the class of simulators considered in this work, behavioral mechanisms are ultimately instantiated as executable programs that can be run, inspected, and modified.
In practice, a simulator is only useful when its rules, interfaces, and data flows are embodied in code that can be run, evaluated, and calibrated~\cite{kennedy2001bayesian,law2007simulation,grimm2020odd}.

However, simulator generation is not merely a code-synthesis problem. 
A generated simulator may compile and produce outputs, yet still implement the wrong behavioral mechanism: an incorrect state-transition rule, a missing interaction pathway, a mis-specified intervention effect, or an output process that matches the required schema but fails to reproduce observed behavior. 
These errors are rarely exposed by code-level checks alone; they often appear as residuals between simulated and observed aggregate behaviors, such as mismatched adoption curves, mobility distributions, transition statistics, or destination-recovery scores. 
This observation motivates a more precise goal: to automate not only the construction of runnable simulator code, but also the diagnosis and repair of simulator mechanisms from behavioral residuals.

Existing pathways only partially meet this bar. 
\emph{Manual, expert-built simulators}~\cite{zhang2025socioverse,DBLP:journals/corr/abs-2502-08691,DBLP:journals/corr/abs-2502-18712,jiawei2024large,yang2024oasis,tang2024gensim,yan2024opencity} can encode rich domain mechanisms but are expensive to design and maintain. 
\emph{Description-driven systems}---for example, YuLan-OneSim~\cite{wang2025yulan}, which converts natural-language scenario descriptions into code using ODD protocol~\cite{grimm2010odd} and behavior graphs---reduce coding burden but do not, by themselves, provide an inner loop for \emph{data calibration} of the generated programs. 
At the other end, recent \emph{automatic frameworks} such as G-Sim~\cite{DBLP:journals/corr/abs-2506-09272} pair LLM-proposed designs with simulation-based inference, showing promising alignment to observed data; however, these efforts have largely focused on aggregate models and may not readily extend to highly heterogeneous, micro-level simulators. 
More importantly, existing systems often leave under-specified the diagnostic step between evaluation and repair: given an aggregate mismatch between simulated and observed behavior, which simulator mechanism should be changed, and how should this change be translated into executable code?

In parallel, \emph{multi-agent code generation} has advanced rapidly: CAMEL~\cite{li2023camel}, ChatDev~\cite{qian2024chatdev}, AgentCoder~\cite{huang2023agentcoder}, AutoGen~\cite{harper2024autogenesisagentselfgeneratingmultiagentsystems}, and AI Scientist-v2~\cite{yamada2025ai} stage specialized LLM agents to propose, test, and refine software or to conduct scientific experiments. 
These frameworks demonstrate compelling task decomposition and collaboration patterns, but they are typically \emph{prompt-sensitive, hand-tuned, and not designed for simulator construction or data calibration}. 
Their agent protocols and success criteria are rarely coupled to behavioral residuals for mechanism-level diagnosis and localized repair.
As a result, they may produce runnable programs without systematically improving the behavioral mechanisms that make a simulator faithful to observed data.

To overcome these challenges, we introduce \textbf{\socia}
(Simulation Orchestration for Computational Intelligence with Agents),
an end-to-end multi-agent framework for diagnosis-driven simulator-code
refinement.
\socia treats simulator building as iterative refinement of executable code
within a \textbf{textual computation graph}.
It defines specialized agents for data analysis, code generation, execution,
evaluation, and feedback, and embeds each agent as a node in a directed graph
that carries textual artifacts, such as code, logs, diagnoses, and repair
feedback, together with numeric signals, such as metrics and residuals.
A centralized workflow manager orchestrates \emph{forward execution}---generate
code, run the simulator, compute validation metrics, and check constraints---and
\emph{backward repair}---attribute behavioral residuals to candidate simulator
mechanisms, identify the spatial or interaction context in which those
mechanisms operate when such grounding is supported by the task and data,
generate localized textual repair signals, and update the affected simulator
components.
For spatially grounded tasks, this attribution is guided by a simulator
blueprint that associates residual-relevant mechanisms with the spatial
entities, transitions, relations, or interaction structures on which they
operate.

Following TextGrad, \socia uses the terminology of \emph{textual gradients}, \emph{Textual Gradient Descent}, \emph{momentum}, and \emph{projected gradient descent (PGD)-style projection}, but operationalizes them as structured code-repair abstractions rather than numerical differentiation. 
A textual gradient denotes a loss-aligned, evidence-grounded repair signal that explains how simulator code should change to reduce downstream behavioral residuals. 
Momentum corresponds to history-aware aggregation of previous diagnoses and patches, helping avoid repeated mistakes during long-horizon editing. 
PGD-style projection corresponds to constraint-aware repair that preserves compilability, interface validity, and output-schema conformance. 
Thus, \socia directly refines executable simulator code, while using metric residuals, grounded mechanism diagnosis, and constraint checking to guide iterative repair.

Our focus on executable simulator code is complementary to learned world models~\cite{hafner2023mastering,bruce2024genie} rather than a claim that code-based simulation is universally preferable. 
Learned world models are particularly attractive when the relevant environment dynamics can be learned reliably from available observational or interaction data and predictive performance is the primary objective~\cite{levy2024learning,chen2025policy}. 
Our setting instead emphasizes simulators with explicit behavioral mechanisms that support intervention and counterfactual analysis, including settings in which the target regime is only partially observed or entirely unseen during construction. 
In such cases, learning expressive agent-specific dynamics directly from data can be difficult, particularly when individual histories are sparse. Executable simulators provide a complementary inductive bias by representing behavioral states, transition rules, interaction pathways, and intervention effects explicitly. 
These mechanisms can be inspected, modified, and re-executed under alternative assumptions. 
The resulting artifact is therefore not only a predictor of future observations, but also a reusable agent-level sandbox for studying interventions and distribution shifts.

In summary, this paper makes three main contributions:
\textbf{(1)} We formulate automated simulator generation as diagnosis-driven iterative refinement, where behavioral residuals are attributed to simulator mechanisms rather than used only as aggregate evaluation scores.
\textbf{(2)} We introduce a spatially grounded residual-to-mechanism formulation in which each candidate mechanism is paired with the spatial entities, transitions, relations, or interaction structures on which it operates. This grounding is propagated through mechanism diagnosis and textual repair, enabling localized structural edits while preserving unrelated simulator components. 
\textbf{(3)} We demonstrate empirically that \socia constructs accurate and extrapolatable simulators on personal mobility and geo-social diffusion benchmarks, covering in-distribution simulation, out-of-distribution what-if generation, individual calibration, and transition dynamics.

%% file: latex/chapter/2_related.tex
\textbf{Multi-agent systems (MAS) in Software Engineering.}
MAS are widely used for code and software generation~\cite{huang2023agentcoder,li2023camel,harper2024autogenesisagentselfgeneratingmultiagentsystems,jin2025llmsllmbasedagentssoftware,islam2025codesim,lin2025soen}. Frameworks like MetaGPT~\cite{DBLP:conf/iclr/HongZCZCWZWYLZR24} and ChatDev~\cite{qian2024chatdev} formalize workflows via structured roles, while CodeR~\cite{chen2024coderissueresolvingmultiagent}, RGD~\cite{jin2024rgdmultillmbasedagent}, UnitTest~\cite{alshahwan2024automatedunittestimprovement}, and AI Scientist-v2~\cite{yamada2025ai} target specific phases (issue resolution, testing, or experimentation). 
In contrast to general software-engineering workflows whose primary objectives are code correctness, testing, or task completion, \socia organizes specialized agents around simulator fidelity: execution results are evaluated against behavioral targets, attributed to simulator mechanisms, and translated into localized simulator repairs.

\textbf{Simulator construction.}
Prior work spans: (1) \textbf{Domain-crafted simulators}, which are manually authored by domain experts and therefore create a substantial engineering bottleneck~\cite{zhang2025socioverse,DBLP:journals/corr/abs-2502-08691,DBLP:journals/corr/abs-2502-18712,jiawei2024large,yang2024oasis,tang2024gensim,yan2024opencity,law2007simulation};
(2) \textbf{LLM-based multi-agent systems} for constructing or operating agent-based simulations~\cite{park2023generativeagentsinteractivesimulacra,DBLP:conf/ijcai/ChenDSZS00024,rasalLLMHarmonyMultiAgent2024,harper2024autogenesisagentselfgeneratingmultiagentsystems,shangAgentSquareAutomaticLLM2025,wang2025yulan,yue2025foam,cui2025talking,du2025cams,liu2026gatsim,shao2026generating,zhang2026autob2g};
(3) \textbf{Automated, data-calibrated pipelines} (e.g., G-Sim) that align models to observations yet emphasize aggregate calibration over localized code repair~\cite{DBLP:journals/corr/abs-2506-09272}.
While parameter calibration can be effective when simulator structure is already adequate, it may be insufficient when behavioral discrepancies arise from missing or misspecified mechanisms.
In contrast, \socia treats executable simulator mechanisms themselves as the object of refinement: metric residuals are attributed to candidate mechanisms through a structured simulator blueprint, and the resulting diagnosis is translated into localized structural code repair rather than only parameter adjustment.
For spatially grounded tasks, the blueprint further associates each candidate mechanism with the spatial entities, transitions, relations, or interaction structures in which it operates, allowing spatial or interaction structure to participate directly in diagnosis and repair.

\textbf{Textual gradients \& iterative self-improvement.}
DSPy~\cite{khattab2024dspy} formalizes LM pipelines as text-transformation graphs and optimizes them algorithmically; 
TextGrad~\cite{DBLP:journals/corr/abs-2406-07496} improves upstream components in compound systems by \emph{backpropagating textual feedback}; 
and Self-Refine~\cite{madaan2023self} and Reflexion~\cite{shinn2023reflexion} demonstrate that iterative critiques and episodic memory outperform one-shot prompting across tasks. 
\socia adopts the textual-gradient abstraction for backward feedback propagation, but differs in how the repair signal is constructed and grounded.
Rather than propagating generic downstream critique, \socia first attributes metric-level behavioral residuals to candidate simulator mechanisms through an explicit blueprint and then converts the resulting diagnosis into localized code-edit instructions.
For spatially grounded tasks, this attribution additionally associates each candidate mechanism with its relevant spatial or interaction context.
This also differs from Reflexion-style self-reflection, where feedback is primarily represented as verbal experience for subsequent attempts: in \socia, feedback is explicitly tied to residual-relevant simulator mechanisms and the contexts in which they operate.
History-aware textual momentum retains previous diagnoses and repairs, while projection preserves executable-code constraints.
Thus, our contribution is not a new numerical gradient formulation, but a simulator-specific, residual-grounded pathway from behavioral mismatch to grounded mechanism attribution and constrained code repair.

%% file: latex/chapter/3_approach.tex
As shown in Figure~\ref{fig:gradient}, \socia addresses a central
difficulty in automated simulator generation: generating runnable code is
not sufficient, because the resulting simulator may still implement the
wrong behavioral mechanisms.
A simulator can compile, execute, and produce outputs, yet fail to capture
the intended agent states, interaction rules, intervention pathways,
transition dynamics, or output-generation logic.
Such mechanism-level errors are often not visible from code-level checks
alone; instead, they appear as residuals between simulated outputs and
observed behavioral signals.

\socia therefore treats simulator construction as
\emph{diagnosis-driven iterative refinement}.
Given a task description and data, \socia generates executable simulator
code, executes the simulator, evaluates its outputs against observed
behavior, diagnoses which simulator mechanisms are likely responsible for
the residual errors, and converts the diagnosis into localized repair
feedback.
This loop is organized as a \emph{textual computation graph}: agents
exchange code, execution traces, metric values, residual summaries,
grounded diagnoses, and repair signals.
Following TextGrad~\cite{DBLP:journals/corr/abs-2406-07496}, we retain
the terminology of \emph{textual gradients}, \emph{Textual Gradient
Descent}, \emph{momentum}, and \emph{projection}; however, in \socia
these terms denote structured credit-assignment and repair abstractions
rather than numerical differentiation.

In this section, \S\ref{ssec:overview} introduces the diagnosis-driven
refinement view; \S\ref{ssec:agent} describes the roles and functionalities of the individual agents;
\S\ref{ssec:diagnosis} formalizes grounded residual-to-mechanism
attribution; and \S\ref{ssec:textgrad} explains how this attribution is propagated through textual-gradient repair, memory, and projection.

\subsection{Diagnosis-Driven Simulator Code Refinement}
\label{ssec:overview}

The key challenge in simulator generation is not merely producing
syntactically valid code, but producing code whose mechanisms match the
target phenomenon.
For example, a generated simulator may contain a plausible state-update
function but use the wrong transition rule; it may include an intervention
variable but connect it to the wrong agent behavior; or it may generate
outputs in the correct schema while failing to reproduce observed
population-level dynamics.
These failures are \emph{mechanism-level} rather than purely syntactic.

\socia closes this gap by converting behavioral residuals into
mechanism-level repair signals.
At iteration \(t\), the current simulator program \(x_t\) is executed on
task inputs \(I\), producing simulator outputs \(o_t\).
The outputs are compared against observed targets \(Y\) using task-specific
metrics, producing a scalar validation objective together with metric-level
residuals.
Rather than treating these residuals only as aggregate scores, \socia uses
them as diagnostic evidence: each residual is linked to plausible simulator
mechanisms and, when supported by the task and data, to the spatial or
interaction structures through which those mechanisms act.

This yields an iterative refinement cycle consisting of simulator generation,
execution, residual evaluation, grounded mechanism diagnosis, localized
repair, and constraint checking.
The update signal is not a numerical gradient; instead, it is a structured
textual repair signal grounded in metric residuals, execution evidence,
mechanism attribution, and constraint information.

\subsection{Orchestrated Agents}
\label{ssec:agent}

\paragraph{\textbf{Workflow Manager}}
The Workflow Manager controls the full refinement loop.
It activates agents, routes artifacts through the textual computation graph,
records iteration history, and terminates refinement when validation
performance plateaus and all hard constraints are satisfied, or the maximum iteration budget is reached.
It also maintains the best feasible simulator encountered so far, allowing
the system to fall back to an earlier valid version when a later repair
introduces failures.

\paragraph{\textbf{Data Analysis Agent (DAA)}}
DAA converts the task description, dataset, and task inputs into a structured simulator blueprint $\mathcal{B}$.
Using a \texttt{data}-\texttt{schema}-\texttt{analysis} prompt, it identifies
data fields, agent attributes, environmental variables, interaction relations,
observable outputs, and links across data sources.
Using a structured task-specification protocol, it then constructs the
simulator brief by specifying:
(1) target phenomena and modeling stance;
(2) temporal resolution, spatial or interaction granularity, and population
size;
(3) agent units, static attributes, and dynamic states;
(4) interaction mechanisms among agents or entities;
(5) information-diffusion or influence pathways when applicable;
(6) external signals or interventions and their access paths to simulator
modules;
(7) actions and decision policies; and
(8) task-relevant metrics.

Crucially, $\mathcal{B}$ additionally records grounding information for later diagnosis.
For each task metric, the blueprint records candidate simulator mechanisms
that may affect that metric and, when supported by the task specification or
data, the spatial or interaction context in which each mechanism operates.
Such context may include spatial entities such as POIs or anchors,
spatiotemporal transitions, pairwise spatial relations, or typed interaction
structures.
The blueprint therefore supports diagnosis of not only \emph{which}
mechanism may explain a residual, but also \emph{through which spatial or
relational structure} that mechanism affects simulator behavior.

\paragraph{\textbf{Code Generation Agent (CGA)}}
CGA is responsible for both initial simulator synthesis and subsequent repair.
It supports three working modes:
\begin{itemize}
    \item \emph{Initial code synthesis.}
    CGA reads the simulator blueprint $\mathcal{B}$ and role description and produces the initial runnable simulator.

    \item \emph{Code refinement.}
    During iterative repair, CGA consumes textual gradients---i.e.,
    structured mechanism-level repair signals---and edits the diagnosed
    simulator mechanisms while preserving unaffected components whenever
    possible.

    \item \emph{Constraint repair.}
    CGA applies minimal edits to satisfy hard constraints such as
    compilability, interface validity, I/O policy, deterministic execution,
    and output-schema conformity.
\end{itemize}
CGA performs a self-check after generation or editing, including
checks for syntax errors, empty functions, interface mismatches, and runtime
risks, followed by localized repair when required.

\paragraph{\textbf{Simulation Execution Agent (SEA)}}
SEA compiles and executes simulator code \(x_t\) on task inputs \(I\).
It returns simulator outputs \(o_t\), intermediate traces when available,
execution logs, timing information, and compilation or runtime errors.
These artifacts form the forward-pass evidence used by later evaluation and
diagnosis.

\paragraph{\textbf{Result Evaluation Agent (REA)}}
REA compares simulator outputs \(o_t\) with observed targets \(Y\) using
task-specific metrics.
Its role is not limited to reporting a scalar objective.
Instead, it preserves metric-level residuals, identifies abnormal output
patterns, and records constraint violations.
REA therefore converts raw simulator behavior into structured evidence that
can be consumed by the diagnosis process.

\paragraph{\textbf{Feedback Generation Agent (FGA)}}
FGA converts residual evidence into mechanism-grounded repair feedback.
Given the current simulator, metric residuals, execution evidence,
constraint violations, and the grounded blueprint produced by DAA, FGA
performs residual-to-mechanism attribution.
For mechanisms with supported spatial or interaction grounding, the
attribution explicitly considers the relevant entities, transitions,
relations, or interaction structures.
The resulting diagnosis identifies one or more likely residual-relevant mechanisms together with the spatial or interaction contexts in which those mechanisms operate.
FGA then converts this grounded diagnosis into a textual gradient:
localized repair feedback specifying which simulator mechanisms should be
revised and why.

Together, these agents separate simulator generation into specification,
execution, evaluation, diagnosis, repair, and constraint checking.
This decomposition is important because aggregate losses alone rarely
indicate how simulator code should change; the intermediate artifacts make
the path from behavioral mismatch to executable repair explicit.

\subsection{From Behavioral Residuals to Grounded Mechanism Diagnosis}
\label{ssec:diagnosis}

The central operation in \socia is to convert metric-level behavioral residuals
into grounded mechanism diagnoses, which are subsequently translated into
localized textual repair feedback.
The residual-to-mechanism principle is task-general, while the grounding is task-dependent. The specific metrics and spatial or interaction structures differ across domains, but the diagnostic principle is the same: observable mismatches are linked to a structured simulator blueprint, plausible responsible mechanisms are identified, and localized repair instructions are produced.

\paragraph{\textbf{Metric Residual Construction}}
Given simulator output \(o_t\) and observed targets \(Y\), REA computes a set
of task-specific metrics:
\[
\boldsymbol{\mu}_t
=
[\mu_t^{(1)},\mu_t^{(2)},\ldots,\mu_t^{(K)}],
\]
where each metric measures one observable dimension of simulator fidelity.
The corresponding residual vector is
\[
\mathbf r_t
=
\mathrm{Residual}
(\boldsymbol{\mu}_t,\boldsymbol{\mu}^{\star}),
\]
where \(\boldsymbol{\mu}^{\star}\) denotes reference metric values derived
from observations.
A scalar validation objective \(L_t\) may aggregate these residuals, but
\socia preserves the individual components of \(\mathbf r_t\) because
different metrics provide evidence about different simulator mechanisms.

\paragraph{\textbf{Grounded Residual-to-Mechanism Linking}}
The simulator blueprint produced by DAA associates each evaluation metric
with candidate simulator mechanisms and, when applicable, their spatial or
interaction grounding:
\begin{equation}
\mathcal B:
\mu^{(k)}
\mapsto
\left\{
(h_j,\rho_j)
\right\}_{j=1}^{q}.
\label{eq:grounded_blueprint}
\end{equation}
Here \(h_j\) denotes a candidate simulator mechanism or code component,
such as an agent state variable, interaction function, transition rule,
intervention handler, calibration parameter, or output-construction routine.
The grounding variable \(\rho_j\) denotes the spatial or interaction structure
through which \(h_j\) influences the corresponding observable, including
spatial entities, spatiotemporal transitions, spatial relations, anchors, or
typed interaction layers.

The grounding is \emph{evidence-bounded}: when no spatial or interaction
structure is supported by the task specification or data, \(\rho_j\) is left
unspecified.
Accordingly, \(\rho_j\) should not be interpreted as a geographic error
location or as a hard spatial constraint.
Instead, it provides structured context for credit assignment by limiting
diagnosis to mechanisms and relational structures that are plausible
explanations of the observed residual.

\paragraph{\textbf{Grounded Mechanism Diagnosis}}
FGA combines the current simulator, metric residuals, grounded blueprint,
execution evidence, and constraint violations to produce a diagnosis:
\begin{equation}
d_t
=
\mathcal D_{\mathrm{LLM}}
(x_t,\mathbf r_t,\mathcal B,e_t,v_t).
\label{eq:grounded_diagnosis}
\end{equation}
The diagnosis may contain multiple grounded mechanism attributions, which we extract as
\begin{equation}
\mathcal A_t
=
\mathrm{Attr}(d_t)
=
\left\{
(\hat h_{t,\ell},\hat\rho_{t,\ell})
\right\}_{\ell=1}^{N_t}.
\label{eq:attribution_set}
\end{equation}
Here \(\hat h_{t,\ell}\) denotes a likely residual-relevant mechanism, while
\(\hat\rho_{t,\ell}\) denotes its associated spatial or interaction context
when such grounding is supported.
The operator \(\mathrm{Attr}(\cdot)\) denotes extraction of the grounded
residual-to-mechanism attributions contained in the diagnosis.
Thus, \(d_t\) is not merely a generic critique: it contains explicit credit
assignment from observed residuals to plausible simulator mechanisms and
their operating contexts.

These attributions are diagnostic rather than causal guarantees.
A residual can be consistent with multiple mechanisms, and the role of
\(\mathcal B\) is to constrain the hypothesis space for repair rather than
to establish causal identification.

\subsection{Textual-Gradient Repair Abstraction}
\label{ssec:textgrad}

Following TextGrad~\cite{DBLP:journals/corr/abs-2406-07496}, we represent
the \socia pipeline as a textual computation graph and use gradient-inspired
terminology to organize backward credit assignment and iterative repair.
These terms are operational abstractions rather than numerical
differentiation: the simulator pipeline is generally non-differentiable, and
we do not assume differentiability or monotonic improvement of the resulting
updates.

\paragraph{\textbf{(1) Problem Definition}}
We study iterative refinement of an executable simulator program.
\begin{itemize}
    \item \textbf{Simulator blueprint:}
    \(\mathcal B\), constructed by DAA from the task description and
    structured data analysis.

    \item \textbf{Data:}
    task inputs \(I\) and observed targets \(Y\).

    \item \textbf{Variable:}
    \(x\in\mathcal X\), a runnable simulator program represented as code.

    \item \textbf{Role description:}
    \(r\), specifying the simulator's intended interface, I/O behavior,
    semantics, and constraints.

    \item \textbf{Hard constraints:}
    \[
    \mathcal C
    =
    \{x:c_j(x)\leq0,\;j=1,\ldots,J\}.
    \]
    These include syntactic validity, compilability, valid file access,
    non-empty execution flow, deterministic behavior when required, and
    output-schema conformity.

    \item \textbf{Refinement objective:}
    \[
    \min_{x\in\mathcal X}L(x)
    \qquad
    \text{s.t.}
    \qquad
    x\in\mathcal C.
    \]
    Here \(L(x)\) measures behavioral mismatch between simulated and
    observed outputs, while \(\mathcal C\) defines executable and
    interface-valid simulator programs.
\end{itemize}
We use \(LLM(z)\) to denote prompting a language model with input \(z\) and
collecting its response.

\paragraph{\textbf{(2) Textual Computation Graph}}
We model the system as a DAG \(\mathcal G=(V,E)\).
Nodes are black-box operators: LLM nodes generate or edit textual artifacts,
while function nodes execute code, compute metrics, check constraints, or
aggregate evidence.
Edges carry simulator code, task specifications, execution outputs, metric
residuals, grounded diagnoses, and repair feedback.

Each node \(v\) computes
\[
z_v
=
f_v(\{z_u:u\in\mathrm{Pa}(v)\}),
\]
where \(\mathrm{Pa}(v)\) denotes the parents of \(v\).
Many \(f_v\) are non-differentiable, including LLM calls, compilers, and
simulators.
We therefore use textual gradients as structured backward signals:
downstream residual evidence is attributed to upstream simulator mechanisms
and converted into localized repair instructions.

\paragraph{\textbf{(3) Forward Pass and Behavioral Residuals}}
Code initialization produces the first simulator:
\[
x_0
=
f_{\mathrm{CGA}}(\mathcal B,r).
\]
At iteration \(t\), SEA executes
\[
o_t
=
f_{\mathrm{Sim}}(x_t,I),
\]
where \(o_t\) contains simulated observables and, when available, execution
traces.
REA computes
\[
L_t
=
\ell(o_t,Y)
+
\lambda
\sum_j
\max(0,c_j(x_t)).
\]
The first term measures behavioral mismatch and the second penalizes
constraint violations.
Importantly, REA also preserves the metric residual vector \(\mathbf r_t\),
execution evidence \(e_t\), and constraint violations \(v_t\), rather than
reducing all downstream evidence to the scalar \(L_t\).

\paragraph{\textbf{(4) Textual Gradient as Grounded Credit Assignment}}
The backward repair signal is represented as
\begin{equation}
g_t
=
\mathcal G_{\mathrm{LLM}}
(x_t,\mathbf r_t,d_t,e_t,v_t),
\label{eq:textual_gradient}
\end{equation}
where \(d_t\) is the grounded diagnosis from
Eq.~\ref{eq:grounded_diagnosis}, and
\(\mathcal A_t=\mathrm{Attr}(d_t)\) denotes its grounded attribution set.



Conceptually, backward responsibility is routed in two stages:
output-level residual evidence is first attributed to plausible simulator
mechanisms, and the resulting grounded diagnosis is then converted into
localized repair feedback. Operationally, the backward pathway is
\[
(\mathbf r_t,\mathcal B,x_t,e_t,v_t)
\rightarrow d_t
\rightarrow \mathcal A_t
\rightarrow g_t .
\]

Consequently, \(g_t\) is neither a numerical gradient vector nor a generic
critique.
It is a localized repair direction conditioned on residual evidence and
grounded mechanism attribution, specifying how diagnosed mechanisms should
be revised while preserving unrelated simulator mechanisms and spatial or
interaction structures whenever possible.

\paragraph{\textbf{(5) Textual Momentum as History-Aware Repair Memory}}
To stabilize long-horizon editing, \socia maintains a textual momentum
buffer:
\[
m_t
=
\mathrm{Decay}_{\beta}(m_{t-1})
\oplus
g_t,
\]
where \(\oplus\) denotes ordered merge and
\(\mathrm{Decay}_{\beta}\) down-weights stale items.
In implementation, \(m_t\) is a concise textual memory over recent diagnoses,
failed repairs, successful fixes, and unresolved issues.
We retain the latest \(K\) iterations, with \(K=3\) in our experiments.

Rather than responding only to the current residual, CGA therefore receives
recent repair context indicating what previously failed, what has already
been repaired, and which simulator mechanisms should remain unchanged.
Textual momentum thus serves as history-aware repair memory within the
gradient-inspired abstraction.

\paragraph{\textbf{(6) Textual Repair and Constraint Projection}}
Given the aggregated textual repair signal \(m_t\), CGA performs the TGD-style
update
\[
x_{t+1}^{\mathrm{unproj}}
=
\mathrm{TGD.step}(x_t,m_t;r),
\]
where \(\mathrm{TGD.step}\) denotes an LLM-based code-repair operator rather
than numerical gradient descent.
It applies the grounded repair signal to diagnosed simulator mechanisms while
preserving unaffected components whenever possible.

A subsequent projection-style repair maps the edited simulator back to the
feasible set:
\[
x_{t+1}
=
\Pi_{\mathcal C}(x_{t+1}^{\mathrm{unproj}}),
\qquad
\Pi_{\mathcal C}:\mathcal X\rightarrow\mathcal C.
\]
The projector performs static checks, interface tests, compilation checks,
schema validation, and other task-specific feasibility checks.
If a violation is detected, CGA performs minimal-edit repair and repeats the
self-check procedure.
The projection step preserves executability and interface validity;
it should not be interpreted as imposing additional geographic or spatial
feasibility constraints.

\paragraph{\textbf{(7) One-Line Summary}}
The complete refinement pathway is:
\begin{simstepbox}
\begin{tabularx}{\linewidth}{@{}l X@{}}

\textbf{Forward:} &
\(
\begin{aligned}[t]
o_t &= f_{\mathrm{Sim}}(x_t,I),\\
L_t &= \ell(o_t,Y)
+\lambda\sum_j\max(0,c_j(x_t)).
\end{aligned}
\)
\\[4pt]

\textbf{Residual:} &
\(
\mathbf r_t
=
\mathrm{Residual}
(\boldsymbol{\mu}_t,\boldsymbol{\mu}^{\star})
\)
\\[4pt]

\textbf{Diagnosis:} &
\(
\begin{aligned}[t]
d_t &=
\mathcal D_{\mathrm{LLM}}
(x_t,\mathbf r_t,\mathcal B,e_t,v_t),\\
\mathcal A_t &= \mathrm{Attr}(d_t).
\end{aligned}
\)
\\[4pt]

\textbf{Backward:} &
\(
g_t
=
\mathcal G_{\mathrm{LLM}}
(x_t,\mathbf r_t,d_t,e_t,v_t)
\)
\\[4pt]

\textbf{Momentum:} &
\(
m_t
=
\mathrm{Decay}_{\beta}(m_{t-1})
\oplus g_t
\)
\\[4pt]

\textbf{Update:} &
\(
\begin{aligned}[t]
x_{t+1}^{\mathrm{unproj}}
&=
\mathrm{TGD.step}(x_t,m_t;r),\\
x_{t+1}
&=
\Pi_{\mathcal C}(x_{t+1}^{\mathrm{unproj}}).
\end{aligned}
\)
\\

\end{tabularx}
\end{simstepbox}

The system stops when validation performance plateaus and all hard constraints are satisfied, or the outer-loop budget
is reached.
Throughout refinement, the Workflow Manager maintains the best feasible
simulator observed so far; if a later repair violates feasibility constraints
or fails to improve validation performance, the earlier incumbent is retained.

\paragraph{\textbf{What the Textual Gradient Encodes}}
The textual gradient in \socia represents three linked operations:
\begin{itemize}
    \item \textbf{Residual-to-grounded diagnosis:}
    metric-level residuals are attributed to plausible simulator mechanisms
    together with their relevant spatial or interaction contexts.

    \item \textbf{Diagnosis-to-localized repair:}
    grounded diagnoses are translated into targeted mechanism-level repair
    instructions while unrelated simulator components are preserved whenever
    possible.

    \item \textbf{Repair-to-feasible simulator:}
    CGA applies the repair signal to executable simulator code, while
    projection preserves compilability, interface validity, output-schema
    conformity, and other hard execution constraints.
\end{itemize}

%% file: latex/chapter/4_exp_setup.tex
\subsection{Benchmarks}
We evaluate \socia on two complementary spatially grounded settings: the real-world LLMob mobility benchmark and a controlled semi-synthetic geo-social mask-adoption benchmark. Together, they cover four evaluation regimes and test \socia across trajectory-level physical mobility simulation and social-spatial information diffusion.

\paragraph{\textbf{(1) Personal Mobility Generation}}
We adopt the LLMob~\cite{jiawei2024large} dataset, which contains real-world spatiotemporal trajectories from residents in Japan. 
The goal is to generate each individual's next-day mobility trajectory from historical observations, requiring the simulator to capture both temporal activity regularities and spatial footprint patterns. 
We evaluate three mobility regimes:

\textbf{Mobility N$\rightarrow$N} is an in-distribution routine-mobility simulation setting, where the simulator is constructed using normal-period history and generates normal-period next-day trajectories.
This regime evaluates whether the generated simulator can recover regular spatiotemporal routines.

\textbf{Mobility A$\rightarrow$A} is an in-distribution abnormal-period simulation setting, where the simulator is constructed using abnormal/pandemic-period history and generates abnormal/pandemic-period next-day trajectories. 
This regime evaluates whether the simulator can capture mobility behavior under shifted but observed conditions.

\textbf{Mobility N$\rightarrow$A} is an out-of-distribution what-if simulation setting, where only normal-period history is available but the simulator must generate abnormal/pandemic-period mobility trajectories. 
This regime is the most challenging setting because it requires the simulator to generalize under distribution shift and to model changes in activity timing, trip frequency, and spatial footprint.
Unlike the two in-distribution regimes, N$\rightarrow$A provides no target-regime trajectory history for direct model fitting, making it a particularly relevant setting for evaluating whether an explicit simulator can combine observed history with prior structural knowledge to support counterfactual generation.

\paragraph{\textbf{(2) Controlled Geo-Social Mask-Adoption Benchmark.}}
We introduce Mask Adoption as a controlled, semi-synthetic benchmark rather than an external real-world dataset. Its generator constructs agents with demographic attributes and daily states indicating mask adoption and information exposure, together with typed family, work/school, and community interaction layers through which public-health information and behavioral influence propagate over time. A public-health intervention is introduced from Day~10 onward, creating a known exogenous shift in the agents' information environment.

Running this predefined generative process produces the reference aggregate adoption curve, individual adoption states, and day-to-day state transitions used as ground-truth observables for evaluation. The controlled construction is intentional: rather than claiming to reproduce a particular real-world population, the benchmark tests whether an automatically generated simulator can recover known geo-social diffusion, heterogeneous interaction, and intervention dynamics at aggregate and individual levels. Inspired by the data paradigms in BESSIE~\cite{DBLP:journals/corr/abs-2203-11414} and pandemic-era mask decision simulators~\cite{mitsopoulos2023masking}, it complements LLMob by providing a controlled mechanism-recovery setting for social-spatial diffusion. The benchmark generator and generated data are released with our code.
The ground-truth generator itself is withheld from \socia and all baselines; they receive only the benchmark inputs and observed training signals defined by the common evaluation protocol.



\subsection{Baseline Methods}
We compare \socia against several baselines. 
\textbf{AI Scientist-v2}~\cite{yamada2025ai} is an automated pipeline that ingests experimental data and constructs/executes ML-based models; however, due to the complexity of simulator construction, AI Scientist-v2 still requires a certain amount of \textit{manual} code adjustment to complete executable simulator generation.
We also include \textbf{YuLan-OneSim}~\cite{wang2025yulan}, which structures scenarios via the ODD protocol~\cite{grimm2010odd}, generates code with a Behavior Graph, and improves it via an iterative \emph{verify--repair--regenerate} loop. 
YuLan-OneSim provides scenario-aware simulation construction, but it is not a general-purpose statistical calibration toolkit. 
For evaluation consistency, we adapt it to the same task inputs, output schemas, and metrics used by the other methods.
The \textbf{G-Sim} family~\cite{DBLP:journals/corr/abs-2506-09272} is another key baseline: it uses LLMs to generate simulator code and integrates gradient-free calibration, including evolutionary strategies (\textbf{G-Sim-ES}) and simulation-based inference (\textbf{G-Sim-SBI}), to produce calibrated, uncertainty-aware simulators.
Finally, we compare with \textbf{Reflexion}~\cite{shinn2023reflexion}, which uses verbal reflections from trial feedback as episodic memory to guide subsequent code updates. 
In our implementation, Reflexion follows the same execution and evaluation protocol as \socia, but replaces \socia's structured feedback-generation and repair process with Reflexion-style self-reflection.

\subsection{Evaluation Metrics}
We report \textbf{means with 95\% confidence intervals}. 
Stochastic LLM- and simulator-based runs are repeated with five seeds~\cite{colas2018many}, while SBI draws $1{,}000$ posterior samples to propagate parameter uncertainty~\cite{falkiewicz2023calibrating}.

For \textbf{LLMob personal mobility simulation}, we evaluate both distribution-level and location-level fidelity. 
Specifically, we use \textbf{JSD} and \textbf{WD} to compare simulated and observed mobility distributions, and \textbf{Top-$k$ POI Recall} to assess whether the simulator recovers key frequently visited locations. 
These metrics jointly measure whether the generated trajectories match the observed mobility distribution and preserve important spatial destinations.

For \textbf{Mask Adoption}, we evaluate adoption dynamics at both aggregate and individual levels. 
\textbf{RMSE} measures the fidelity of the global mask-adoption curve over time. 
\textbf{Brier score} evaluates individual-level probabilistic calibration of mask-wearing decisions. 
\textbf{Transition Divergence (TD)} measures how well the simulator reproduces day-to-day adoption-state transitions under information exposure and social influence. 
Unless otherwise specified, lower values indicate better performance, except for Top-$k$ POI Recall where higher is better.

\subsection{Implementation and Refinement Budget}
Unless otherwise specified, all \socia agents, and the reproduced baselines, are built on GPT-5.2~\citep{openai2025gpt5}. 
We re-implemented baselines either via code reproduction or via dataset-level reproduction to ensure evaluation consistency.
%
Across tasks, \socia typically requires 3--5 outer-loop iterations to obtain stable, high-quality simulator code. In our setup, the first simulator draft is generated in approximately 20 minutes, and each subsequent refinement takes approximately 10 minutes, resulting in a total end-to-end wall-clock time of roughly 40--60 minutes per task. Most of this latency comes from LLM inference and simulator execution: code generation is the dominant LLM call, while diagnosis and patch-generation calls are shorter; simulator runtime varies with the complexity of the generated model and any task-specific calibration it invokes.

Each iteration makes a fixed number of LLM calls involving the data analysis, code generation, execution and evaluation, and feedback generation agents, yielding an overall budget of approximately $10^6$ tokens per task and an API cost of approximately \$1.50--\$2.00 in our setup. Because the compared methods use different internal refinement and stopping procedures, we preserve their native algorithms rather than equating iteration counts across methods; all reproduced LLM-based baselines use the same backbone and common task inputs, output schemas, and evaluation protocol.

The additional computation of \socia is most justified when simulator structure itself may be misspecified, or when OOD what-if generation, mechanism inspection, and reusable intervention analysis are important. When the simulator structure is already trusted and the objective is primarily low-cost in-distribution calibration, calibration-only approaches such as G-Sim can be computationally more attractive. We stop \socia when validation performance plateaus and all hard constraints are satisfied, or when the maximum outer-loop budget is reached.

%% file: latex/chapter/5_exp_evaluation.tex
\input{latex/table/llmob_comparison}
\subsection{Simulator Overview}
\label{ssec:simulators}
\socia produces task-specific executable simulators with mechanisms consistent with the corresponding spatial setting. For Personal Mobility, the generated simulator models individual activity schedules, home/work anchors, destination selection, and spatiotemporal travel constraints. For Mask Adoption, it models individual behavioral states and information diffusion over typed family, work/school, and community interaction layers. Detailed simulator structures and representative generated mechanisms are provided in Appendix~\ref{appen:simulator_examples}.

\subsection{Findings (Tables~\ref{tab:overall-results} and~\ref{tab:mask-results})}
\label{ssec:main_findings}

\input{latex/table/mask_comparison}

\paragraph{\textbf{(1) \socia achieves the strongest overall mobility simulation performance.}}
Table~\ref{tab:overall-results} shows that \socia obtains the best result on seven out of nine LLMob metrics across the three mobility regimes. 
For normal$\rightarrow$normal simulation, \socia achieves the lowest JSD ($0.041$) and the highest POI recall ($0.811$), while remaining second-best on WD. 
For abnormal$\rightarrow$abnormal simulation, \socia again performs best on JSD ($0.007$) and POI recall ($0.997$), and second-best on WD. 
Most importantly, in the OOD normal$\rightarrow$abnormal setting, \socia achieves the best scores on all three metrics: JSD ($0.043$), WD ($0.289$), and POI recall ($0.614$). 
This indicates that the generated simulator not only fits in-distribution mobility patterns but also better supports what-if generation under distribution shift.

\paragraph{\textbf{(2) Distributional shape and destination recovery favor mechanism-level repair.}}
JSD and WD capture complementary aspects of mobility fidelity: JSD measures distribution-shape divergence, whereas WD measures distributional displacement.
The results show that \socia is strong in matching the shape of observed mobility distributions, with large improvements over AI Scientist-v2, YuLan-OneSim, Reflexion, and G-Sim in most regimes. 
At the same time, Top-$k$ POI Recall shows that \socia also recovers key frequently visited locations more reliably. 
This is particularly clear in the OOD N$\rightarrow$A setting, where \socia reaches $0.614$ POI recall, substantially higher than G-Sim-ES ($0.105$), G-Sim-SBI ($0.214$), and Reflexion ($0.220$). 
Thus, \socia does not merely reduce an aggregate discrepancy; it improves simulator mechanisms related to distributional mobility structure and destination-selection fidelity.

\paragraph{\textbf{(3) \socia also leads on geo-social mask-adoption diffusion.}}
Table~\ref{tab:mask-results} shows that \socia achieves the best performance on all three Mask Adoption metrics. 
It obtains the lowest aggregate adoption RMSE ($0.075$), outperforming the strongest competing baseline, G-Sim-ES ($0.089$). 
It also achieves the best Brier score ($0.255$), indicating better individual-level probabilistic calibration of mask-wearing decisions. 
Finally, \socia obtains the lowest Transition Divergence ($0.161$), showing that it better reproduces day-to-day adoption-state transitions under information exposure and social influence. 
These results suggest that the generated simulator captures not only the global adoption curve but also the individual decision and temporal transition mechanisms underlying geo-social diffusion.


\paragraph{\textbf{(4) Comparison with G-Sim reveals a metric- and setting-dependent trade-off between parameter calibration and structural repair.} }
G-Sim remains a strong calibration-oriented baseline. 
In particular, G-Sim-ES achieves the lowest WD in the two in-distribution mobility regimes: 0.225 versus 0.332 for \socia in N$\rightarrow$N, and 0.467 versus 0.532 in A$\rightarrow$A. This indicates that when target-regime observations are available, parameter calibration can be highly effective at reducing distributional displacement even without modifying simulator structure. However, these same G-Sim-ES solutions have substantially larger JSD than \socia (0.546 vs.\ 0.041 in N$\rightarrow$N and 0.272 vs.\ 0.007 in A$\rightarrow$A), showing that lower WD does not necessarily imply faithful recovery of the overall distributional shape.  

We also observe that, under our reproduced LLMob setting, G-Sim-ES obtains lower WD than G-Sim-SBI, whereas the original G-Sim study reports stronger SBI performance in some of its settings. Our results are obtained under a unified task protocol, backbone, simulator construction setting, and evaluation metrics rather than the original G-Sim experimental configuration; therefore, the relative ordering of ES and SBI should be interpreted as setting- and metric-dependent rather than as general dominance of one calibration procedure. Indeed, G-Sim-SBI remains highly competitive on JSD in A$\rightarrow$A and N$\rightarrow$A, despite its weaker WD.  

The distinction becomes clearest under the OOD N$\rightarrow$A setting, where \socia achieves the best JSD, WD, and POI Recall simultaneously. Overall, these results suggest that calibration-only methods can be highly competitive on individual metrics in observed regimes, whereas mechanism-level repair becomes particularly useful when fidelity must be maintained jointly across metrics or under distribution shift.

\paragraph{\textbf{(5) Comparison with Reflexion shows the benefit of loss-aligned localized repair.}}
Reflexion improves over more general automated-generation baselines, but it remains weaker than \socia across both LLMob and Mask Adoption. 
The gap is especially visible in the OOD mobility setting and in the Mask transition metric. 
This supports the role of \socia's loss-aligned repair loop: evaluation residuals, execution failures, and constraint violations are converted into localized code-edit signals rather than generic self-reflection. 
As a result, \socia can preserve working simulator components while repairing specific mechanisms such as mobility conditioning, POI selection, information exposure, and adoption-state transitions.

\paragraph{\textbf{(6) Why AI Scientist-v2 and YuLan-OneSim trail.}}
AI Scientist-v2 is designed primarily for automated ML experimentation rather than simulator construction, and therefore lacks a dedicated loop for generating, executing, evaluating, and repairing simulator code against task-specific behavioral metrics. 
YuLan-OneSim provides useful scenario structuring and behavior-graph generation, but it lacks loss-aligned, instance-level code refinement.
Consequently, both methods perform worse across the LLMob mobility regimes and the Mask Adoption task. 
The results indicate that simulator generation requires more than executable code synthesis: it requires continuous calibration against observed behavior, structured diagnostics, and mechanism-level repair.

\paragraph{\textbf{(7) Method-level interpretation.}}
The results directly support the core design of \socia. 
First, improvements across multiple metric types show that \socia does not merely target a single aggregate score: it improves distributional fidelity, destination recovery, individual calibration, and transition dynamics.
Second, the comparison with G-Sim shows that parameter calibration can be highly competitive on individual in-distribution metrics, while mechanism-level repair provides more consistent gains when fidelity must be maintained jointly across metrics or under distribution shift.
Third, the comparison with Reflexion shows that generic self-reflection is weaker than residual-grounded, localized repair. 
Together, these findings validate the residual-to-diagnosis and diagnosis-to-repair design: different metric residuals expose different simulator mechanisms, and \socia uses structured feedback to repair them jointly.
A representative residual-to-diagnosis-to-repair trace is provided in Appendix~\ref{appen:diagnosis_case}.

\paragraph{\textbf{Takeaway.}}
Across trajectory-level physical mobility and geo-social information diffusion, \socia delivers the strongest overall simulator fidelity. 
Its advantage is most pronounced when the task requires both distributional alignment and structural reasoning, such as OOD mobility generation and mask-adoption transition dynamics. 
These findings support the central claim that loss-aligned, constraint-aware, and localized simulator-code repair can produce more faithful executable simulators than one-shot generation, generic reflection, or calibration-only baselines.

\subsection{Ablation and Backbone Sensitivity (Table~\ref{tab:ablation_study})}
\label{ssec:ablation}

We conduct an ablation and backbone-sensitivity study to quantify the
contribution of each component in \socia. Unless otherwise noted,
we report performance deltas ($\Delta$) relative to the full GPT-5.2-based
\socia configuration on Mask Adoption and Personal Mobility
N$\rightarrow$A. For Mask Adoption, we report $\Delta$RMSE; for Personal
Mobility N$\rightarrow$A, we report $\Delta$JSD and $\Delta$WD. Positive
values indicate performance degradation, while negative values indicate
improvement over the full configuration.

The ablation variants are:
\begin{itemize}
    \item \textbf{w/o mom.:} remove the history-aware repair memory,
    i.e., the buffer that summarizes previous critiques and code patches.

    \item \textbf{w/o proj.:} remove the projection-style constraint
    repair step, so generated code is no longer explicitly repaired
    against compilation, interface, schema, and other execution
    constraints before the next iteration.

    \item \textbf{w/o data-analysis:} replace the Data Analysis Agent
    with a naive data-understanding prompt, thereby weakening structured
    schema analysis, task specification, metric-to-mechanism linking,
    and task-specific data-to-simulator grounding.

    \item \textbf{w/o eval.:} replace the Result Evaluation Agent with
    a heuristic diagnostic prompt that only asks the LLM to infer code
    problems from the computed metrics, without structured residual
    analysis or component-level error localization.

    \item \textbf{w/o feedback:} replace the Feedback Generation Agent
    with a coarse feedback prompt that asks the LLM to suggest code
    improvements from the current diagnostics, without structured
    attribution, localized repair proposals, or feedback synthesis.

    \item \textbf{w/o iter:} disable iterative code evolution and use
    only the first simulator generated by the code-generation agent.
\end{itemize}

In addition to component ablations, we replace GPT-5.2 with
Qwen3-Next-80B-A3B-Instruct and Llama-3.3-70B-Instruct-Turbo while
keeping the task settings, evaluation protocol, and iterative refinement
pipeline unchanged.

\textbf{Results.}
Table~\ref{tab:ablation_study} shows that removing any major \socia
component degrades performance on both Mask Adoption and Mobility
N$\rightarrow$A. The largest degradation occurs without iterative
refinement: w/o iter increases Mask RMSE by $+0.413$, Mobility JSD by
$+0.317$, and Mobility WD by $+0.370$. Among the agent-role ablations,
w/o feedback causes the largest degradation, followed by w/o eval. and
w/o data-analysis. Removing momentum or projection produces smaller but
consistent performance drops.

The backbone results are mixed. Qwen3-Next-80B remains competitive with
the GPT-5.2 configuration, with $\Delta$RMSE $=-0.024$,
$\Delta$JSD $=+0.002$, and $\Delta$WD $=-0.193$. In contrast,
Llama-3.3-70B substantially degrades all three metrics, with
$\Delta$RMSE $=+0.666$, $\Delta$JSD $=+0.506$, and
$\Delta$WD $=+1.494$. Detailed component-wise interpretations are
provided in Appendix~\ref{appen:ablation_details}.

\subsection{Additional Refinement Stability Analysis}
\label{ssec:supple_exp}

\input{latex/table/ablation_study}

Appendix~\ref{appen:add_exps} reports refinement trajectories over outer-loop iterations, and it shows that \socia reaches a stable low-error region within a few iterations, whereas G-Sim-SBI shows stronger oscillations across Mask Adoption and OOD Mobility metrics. 
This further supports the role of residual-grounded diagnosis and repair memory in stabilizing iterative simulator-code refinement.

%% file: latex/table/llmob_comparison.tex
\begin{table*}[t]
  \centering
  \small
 \caption{
Evaluation results ($\pm$ denotes 95\% CIs) on the \textbf{LLMob personal mobility simulation} benchmark.
Mob. N$\rightarrow$N and Mob. A$\rightarrow$A are in-distribution settings, while Mob. N$\rightarrow$A is an out-of-distribution what-if setting.
Lower is better for JSD/WD and higher is better for POI.
Best and second-best results are shown in \textbf{bold} and \underline{underlined}, respectively.
}
  \label{tab:overall-results}
  \resizebox{\textwidth}{!}{%
  \begin{tabular}{l||lll|lll|lll}
    \hline
    \multirow{2}{*}{Methods $\downarrow$} & \multicolumn{3}{c|}{Mob. N$\rightarrow$N (ID)} & \multicolumn{3}{c|}{Mob. A$\rightarrow$A (ID)}  & \multicolumn{3}{c}{Mob. N$\rightarrow$A (OOD)}  \\
    \cline{2-10} 
      & JSD $\downarrow$ & WD $\downarrow$ & POI $\uparrow$ & JSD $\downarrow$ & WD $\downarrow$ & POI $\uparrow$ & JSD $\downarrow$ & WD $\downarrow$ & POI $\uparrow$\\ \hline
    AI Scientist-v2   & 0.887 $\pm$ 0.006   & 3.224 $\pm$ 0.006 & 0.003 $\pm$ 0.001 & 0.451 $\pm$ 0.006 & 3.373 $\pm$ 0.006 & 0.190 $\pm$ 0.003	 & 0.773 $\pm$ 0.007 & 2.045 $\pm$ 0.006 & 0.070 $\pm$ 0.007 \\
    YuLan-OneSim  & 0.793 $\pm$ 0.004 & 2.191 $\pm$ 0.004 & 0.200 $\pm$ 0.008 & 0.366 $\pm$ 0.005 & 2.485 $\pm$ 0.005 & 0.310 $\pm$ 0.004 & 0.639 $\pm$ 0.006 & 1.541 $\pm$ 0.006 & 0.091 $\pm$ 0.008 \\ 
    Reflexion & \underline{0.423 $\pm$ 0.005} & 0.783 $\pm$ 0.006 & 0.410 $\pm$ 0.006 & 0.187 $\pm$ 0.006 & 0.713 $\pm$ 0.006 & 0.550 $\pm$ 0.006 & 0.553 $\pm$ 0.007 & 0.612 $\pm$ 0.007 & \underline{0.220 $\pm$ 0.003}  \\
    G-Sim-ES  & 	0.546 $\pm$ 0.002  & \textbf{0.225 $\pm$ 0.003} & \underline{0.802 $\pm$ 0.002} & 	0.272 $\pm$ 0.004  & \textbf{0.467 $\pm$ 0.004} 	& 0.601 $\pm$ 0.002 & 0.534 $\pm$ 0.005  	& \underline{0.456 $\pm$ 0.005}  & 0.105 $\pm$ 0.002 \\
    G-Sim-SBI  & 0.635 $\pm$ 0.003  & 1.638 $\pm$ 0.003 & 0.097 $\pm$ 0.001 & \underline{0.012 $\pm$ 0.003}	 & 1.690 $\pm$ 0.006  & \underline{0.990 $\pm$ 0.002} & \underline{0.045 $\pm$ 0.005} & 0.633 $\pm$ 0.003  & 0.214 $\pm$ 0.004 \\
     \hline
    \rowcolor{orange!15}
    \socia  & \textbf{0.041 $\pm$ 0.002} & \underline{0.332 $\pm$ 0.003} & \textbf{0.811 $\pm$ 0.002} & \textbf{0.007 $\pm$ 0.001 }	& \underline{0.532 $\pm$ 0.003 } & \textbf{0.997 $\pm$ 0.001} & \textbf{0.043 $\pm$ 0.005} & \textbf{0.289 $\pm$ 0.006}  & \textbf{0.614 $\pm$ 0.003} \\ 
    \hline
  \end{tabular}
  }
\end{table*}

%% file: latex/table/mask_comparison.tex
\begin{table}[t]
  \centering
  \small
    \caption{Evaluation results ($\pm$ denotes 95\% CIs) on the \textbf{Mask Adoption} task. \textbf{RMSE} measures global adoption-curve fidelity, \textbf{Brier} evaluates individual-level probabilistic calibration of mask-wearing decisions, and \textbf{TD} (Transition Divergence) measures day-to-day adoption-state transition fidelity under information exposure and social influence. Lower values indicate better performance.}
  \label{tab:mask-results}
  \begin{tabular}{lccc}
    \hline
    Method & RMSE $\downarrow$ & Brier $\downarrow$ & TD $\downarrow$ \\ 
    \hline
    AI Scientist-v2 & $0.551 \pm 0.010$ & $0.424 \pm 0.002$ & $0.523 \pm 0.004$ \\
    YuLan-OneSim    & $0.413 \pm 0.009$ & $0.391 \pm 0.006$ & $0.410 \pm 0.004$ \\ 
    Reflexion       & $0.263 \pm 0.005$ & $0.334 \pm 0.003$ & $0.321 \pm 0.005$ \\
    G-Sim-ES        & \underline{$0.089 \pm 0.002$} & 0.298 $\pm$ 0.003 & $0.279 \pm 0.002$ \\
    G-Sim-SBI       & $0.199 \pm 0.003$ & \underline{$0.285 \pm 0.003$} & \underline{$0.194 \pm 0.002$} \\
    \hline
    \rowcolor{orange!15}
    \socia          & \textbf{0.075 $\pm$ 0.007} & \textbf{0.255 $\pm$ 0.002} & \textbf{0.161 $\pm$ 0.007} \\ 
    \hline
  \end{tabular}
\end{table}

%% file: latex/table/ablation_study.tex
\begin{table}[t]
  \centering
  \small
  \caption{Ablation and backbone-sensitivity results on \textbf{Mask Adoption}
and \textbf{Personal Mobility N$\rightarrow$A}. Deltas are relative to the
full GPT-5.2-based \socia configuration; positive values indicate degradation
and negative values indicate improvement.}
  \label{tab:ablation_study}

  \setlength{\tabcolsep}{5.5pt}
  \renewcommand{\arraystretch}{1.08}

  \begin{tabular}{lccc}
    \toprule
    \multirow{2}{*}{\textbf{Models} $\downarrow$}
      & \textbf{Mask.}
      & \multicolumn{2}{c}{\textbf{Mob. N}$\rightarrow$\textbf{A}} \\
    \cmidrule(lr){2-2}
    \cmidrule(lr){3-4}
      & $\Delta$RMSE $\downarrow$
      & $\Delta$JSD $\downarrow$
      & $\Delta$WD $\downarrow$ \\
    \midrule

    \rowcolor{orange!15}
    \socia
      & -
      & -
      & - \\

    \midrule

    w/o mom.
      & \pos{+0.123 $\pm$ 0.003}
      & \pos{+0.056 $\pm$ 0.002}
      & \pos{+0.093 $\pm$ 0.006} \\

    w/o proj.
      & \pos{+0.173 $\pm$ 0.003}
      & \pos{+0.098 $\pm$ 0.005}
      & \pos{+0.131 $\pm$ 0.006} \\

    w/o data-analysis
      & \pos{+0.307 $\pm$ 0.005}
      & \pos{+0.203 $\pm$ 0.008}
      & \pos{+0.247 $\pm$ 0.003} \\

    w/o eval.
      & \pos{+0.362 $\pm$ 0.004}
      & \pos{+0.207 $\pm$ 0.006}
      & \pos{+0.235 $\pm$ 0.004} \\

    w/o feedback
      & \pos{+0.381 $\pm$ 0.004}
      & \pos{+0.264 $\pm$ 0.006}
      & \pos{+0.314 $\pm$ 0.007} \\

    w/o iter
      & \pos{+0.413 $\pm$ 0.007}
      & \pos{+0.317 $\pm$ 0.009}
      & \pos{+0.370 $\pm$ 0.004} \\

    \midrule

    Qwen3-Next-80B
      & \negv{-0.024 $\pm$ 0.003}
      & \pos{+0.002 $\pm$ 0.001}
      & \negv{-0.193 $\pm$ 0.003} \\

    Llama-3.3-70B
      & \pos{+0.666 $\pm$ 0.007}
      & \pos{+0.506 $\pm$ 0.006}
      & \pos{+1.494 $\pm$ 0.004} \\

    \bottomrule
  \end{tabular}
\end{table}

%% file: latex/chapter/6_conclusion.tex
\paragraph{\textbf{Conclusion.}}
We introduced \textbf{\socia}, a diagnosis-driven multi-agent framework for automated simulator-code construction. 
\socia directly refines executable simulator code through execution, evaluation, grounded mechanism diagnosis, localized textual repair, history-aware memory, and constraint projection.
Across the real-world LLMob mobility benchmark and the controlled semi-synthetic Mask Adoption benchmark, \socia achieves the strongest overall simulator fidelity, including robust OOD mobility generation and better aggregate, individual-level, and transition-level mask-adoption dynamics. 
Ablations further show that iteration, structured data analysis, evaluation diagnostics, feedback generation, repair memory, and constraint projection are all important for high-fidelity simulator construction.

\paragraph{\textbf{Limitations.}}
The present evaluation considers simulators of moderate scale and complexity, and \ofsocia cost may increase with population size, interaction density, and calibration dimensionality. We also do not yet evaluate region-specific errors or calibration disparities across demographic subgroups, which are important for spatially heterogeneous deployments. Generated mechanisms and code should therefore remain subject to human review in high-stakes settings. In public-health applications, \socia is intended for decision support and scenario analysis rather than autonomous decision making.

\paragraph{\textbf{Future Work.}}
We do not claim that executable simulator construction dominates learned world models in fully supervised, in-distribution settings. A systematic comparison between code-based simulators and learned dynamics models across different data regimes is an important direction for future work, particularly to characterize when editability and mechanism transparency outweigh the predictive efficiency of purely learned models. We also plan to evaluate \socia on larger and more complex simulators and to extend its diagnostic pipeline toward region- and subgroup-specific residual analysis for spatially heterogeneous deployments.

%% file: latex/chapter/acknow.tex
This research is supported by the ARC Center of Excellence for Automated Decision Making and Society (CE200100005).
Computational facilities were provided by the School of Computer Science and Engineering at UNSW Sydney through the Wolfpack computational cluster. 
We used icon assets from \textit{Flaticon.com} to create the illustrations in this paper. We appreciate the artists who created and shared these icons on \url{Flaticon.com}.

%% file: latex/supplementary/simulator_overview.tex
Our \socia derives task-specific simulator code through data-driven execution, evaluation, and repair. 
We focus on two complementary evaluation settings: the real-world \textbf{LLMob personal-mobility benchmark}, which evaluates trajectory-level physical mobility, and the controlled semi-synthetic \textbf{Mask Adoption benchmark}, which evaluates geo-social information diffusion over typed interaction spaces. 
By inspecting the generated code, we briefly summarize the simulator structures learned by \socia.

\paragraph{\textbf{Personal Mobility Generation.}} 
\socia builds the Personal Mobility Simulator as a city-scale, agent-based pipeline that consumes an individual's historical spatiotemporal traces and generates a feasible next-day itinerary with timestamps and locations. 
The simulator defines \emph{Resident}, \emph{Location}, and \emph{Simulation Environment} primitives, and derives resident profiles, including home/work anchors, daily routines, activity preferences, and temporal activity patterns. 
An LLM-based schedule generator then plans a day as a sequence of activities, such as work, dining, study, and shopping, and selects candidate POIs using preference- and context-aware scoring under spatiotemporal constraints such as opening hours, travel-time budgets, and inter-POI distance. 
The same simulator structure supports three regimes by switching conditioning data and behavioral priors: normal$\rightarrow$normal, abnormal$\rightarrow$abnormal, and normal$\rightarrow$abnormal. 
The last setting is an OOD what-if setting, requiring the simulator to generate abnormal/pandemic trajectories from normal-period history.

\paragraph{\textbf{Mask Adoption.}}
\socia implements the Mask Adoption Simulator as a geo-social diffusion model over typed interaction spaces. 
Although this task is not a geographic trajectory-generation task, it is spatially relevant in the geo-social sense: agents are embedded in family, work/school, and community interaction layers, where information exposure and behavioral influence propagate over time. 
Each agent maintains dynamic states such as mask status, risk perception, information exposure, and behavioral memory. 
At each daily step, the simulator computes a latent adoption score by combining personal profile factors, multi-layer neighborhood influence, temporal persistence, and intervention-related information exposure; this score is mapped to a probabilistic mask-wearing decision. 
The public-health information shock around Day~10 is modeled as an external intervention whose effect propagates through the typed interaction layers, allowing the simulator to capture not only aggregate adoption trends but also individual decision calibration and day-to-day behavioral transitions. 
This benchmark therefore complements LLMob by evaluating \socia on social-spatial diffusion rather than physical mobility.

%% file: latex/supplementary/ablation_details.tex
This section provides additional interpretation of the ablation results
reported in Table~\ref{tab:ablation_study}.

\paragraph{\textbf{History-aware repair memory.}}
Removing the memory buffer causes moderate but consistent degradation
across both tasks. This suggests that retaining recent diagnoses and
repairs helps prevent repeated mistakes and reduces the tendency to undo
useful previous edits.

\paragraph{\textbf{Constraint projection.}}
Removing projection also degrades all reported metrics. The result
supports the role of explicit repair against compilation, interface,
schema, and execution constraints: localized behavioral repair is more
effective when candidate code updates are prevented from introducing new
feasibility failures.

\paragraph{\textbf{Data Analysis Agent.}}
Replacing DAA with a naive data-analysis prompt substantially weakens
performance. DAA does more than summarize the dataset: it constructs the
structured simulator blueprint that links metrics to candidate mechanisms
and, when supported by the task and data, to their spatial or interaction
grounding. Weakening this stage therefore reduces the quality of later
residual-to-mechanism attribution.

\paragraph{\textbf{Result Evaluation Agent.}}
Replacing REA with heuristic diagnostics causes substantial degradation,
showing that scalar losses alone provide insufficient information for
repair. Structured metric-level residuals and execution evidence provide
more informative signals for localizing simulator failures.

\paragraph{\textbf{Feedback Generation Agent.}}
Removing structured FGA produces the largest degradation among the
agent-role ablations. This indicates that effective simulator repair
depends not only on observing residuals but also on translating them into
mechanism-level, localized repair instructions. Generic improvement
suggestions are less effective at identifying which simulator mechanisms
should be revised.

\paragraph{\textbf{Iterative refinement.}}
The largest overall degradation occurs when iterative refinement is
disabled. This supports the central design of \socia: one-shot
simulator generation is insufficient, especially under distribution shift,
and repeated execution, evaluation, diagnosis, and repair are necessary
to obtain high-fidelity simulator behavior.

\paragraph{\textbf{Backbone sensitivity.}}
The Qwen3-Next-80B results show that the \socia pipeline is not
intrinsically tied to GPT-5.2. However, the large degradation with
Llama-3.3-70B shows that the framework still depends on reliable
instruction following, diagnosis, and code generation at individual LLM
nodes. A weak backbone can therefore disrupt either the
residual-to-diagnosis or diagnosis-to-repair stage.

%% file: latex/supplementary/0_supple.tex
\input{latex/figure/supple}

Figure~\ref{fig:metric_trends} compares the refinement trajectories of \socia and G-Sim-SBI over outer-loop iterations under the same calibration setting. 
We track three representative macro-level metrics: Mask Adoption RMSE, Mobility N$\rightarrow$A JSD, and Mobility N$\rightarrow$A WD. 
These metrics cover both geo-social diffusion fidelity and out-of-distribution mobility fidelity.

The main observation is that \socia reaches a stable low-error region within the first few iterations. 
For Mask Adoption, RMSE decreases consistently from $0.191$ at iteration 0 to $0.075$ at iteration 3, and then remains close to this low-error region in later iterations. 
For Mobility N$\rightarrow$A, JSD also drops from $0.109$ to $0.043$ by iteration 3, while WD decreases from $0.581$ to $0.289$. 
Although later iterations are not strictly monotonic, the fluctuations are relatively small compared with the initial improvement, suggesting that \socia quickly identifies and repairs the dominant simulator-design errors.

In contrast, G-Sim-SBI exhibits more pronounced oscillations. 
For Mask Adoption, RMSE alternates between high and low values across iterations, e.g., $0.436$ at iteration 1, $0.217$ at iteration 2, $0.365$ at iteration 3, and $0.199$ at iteration 5. 
For Mobility N$\rightarrow$A, G-Sim-SBI sometimes improves one metric but then regresses in later iterations, especially on WD, which drops from $2.15$ to $0.633$ by iteration 3 but rises again to $1.226$ at iteration 4 and $1.000$ at iteration 6. 
This indicates that calibration can reduce a metric at one iteration without necessarily maintaining that improvement across subsequent iterations.

These trajectories support the design motivation of \socia. 
Instead of only searching for better calibrated parameters, \socia converts metric residuals into mechanism-level diagnoses and localized code repairs. 
The history-aware momentum buffer helps the system remember previous failures and avoid repeatedly undoing useful repairs, while PGD-style projection prevents code edits from violating hard execution and schema constraints. 
As a result, \socia typically reaches a stable, high-fidelity region after a small number of iterations, whereas G-Sim-SBI is more vulnerable to oscillatory behavior across metrics.

%% file: latex/figure/supple.tex
\pgfplotstableread{
iter mask_rmse na_jsd na_wd mask_rmse1 na_jsd1 na_wd1
0    0.191     0.109    0.581    0.388     0.05     2.15
1    0.166     0.113    0.481    0.436     0.02     1.23
2    0.124     0.061    0.367    0.217     0.0168   0.9457
3    0.075     0.043    0.289     0.365     0.045   0.6328
4    0.083     0.078    0.299    0.298     0.015    1.226
5    0.082     0.105    0.306    0.199     0.0127   0.8025
6    0.079     0.056    0.323    0.373     0.0437   1.000
}\datatable

\begin{figure*}[t]
\centering
\begin{tikzpicture}

\begin{groupplot}[
  group style={
    group size=3 by 2,
    horizontal sep=1.35cm,
    vertical sep=1.25cm
  },
  width=0.225\textwidth,
  height=3.25cm,
  xmin=0, xmax=6,
  xtick={0,1,2,3,4,5,6},
  xlabel={},
  ylabel={},
  grid=both,
  tick label style={font=\scriptsize},
  label style={font=\scriptsize},
  title style={font=\scriptsize},
  enlarge x limits=0.02,
  enlarge y limits=0.08,
  scaled y ticks=false,
  yticklabel style={
    /pgf/number format/fixed,
    /pgf/number format/precision=3
  },
  every axis plot/.append style={
    mark=*,
    mark size=1.6pt,
    line width=0.8pt
  },
]

\nextgroupplot[title={Mask. RMSE}]
\addplot table[x=iter,y=mask_rmse] {\datatable};

\nextgroupplot[title={Mob. N$\to$A JSD}]
\addplot table[x=iter,y=na_jsd] {\datatable};

\nextgroupplot[title={Mob. N$\to$A WD}]
\addplot table[x=iter,y=na_wd] {\datatable};

\nextgroupplot[title={Mask. RMSE}]
\addplot[red] table[x=iter,y=mask_rmse1] {\datatable};

\nextgroupplot[title={Mob. N$\to$A JSD}]
\addplot[red] table[x=iter,y=na_jsd1] {\datatable};

\nextgroupplot[title={Mob. N$\to$A WD}]
\addplot[red] table[x=iter,y=na_wd1] {\datatable};

\end{groupplot}

\node[font=\scriptsize\bfseries, anchor=east] at ($(group c1r1.west)+(-0.75cm,0)$) {\socia};
\node[font=\scriptsize\bfseries, anchor=east] at ($(group c1r2.west)+(-0.75cm,0)$) {G-Sim-SBI};

\end{tikzpicture}

\caption{Evidence-based refinement trajectories over outer-loop iterations. We compare \socia and G-Sim-SBI under the same calibration setting to examine how each method improves simulator fidelity during iterative refinement. The top row shows \socia, and the bottom row shows G-Sim-SBI. Across columns, we track three macro-level metrics: Mask Adoption RMSE, and OOD Mobility (N$\rightarrow$A) measured by JSD and Wasserstein distance (WD). Lower values indicate better performance.}
\label{fig:metric_trends}
\end{figure*}

%% file: latex/supplementary/2_case_study.tex
To examine whether the diagnosis module identifies meaningful simulator-level failure mechanisms, we present a representative repair trace from the LLMob N$\rightarrow$N experiment. The case shows how a behavioral residual is translated into a mechanism-level diagnosis and a localized structural modification.

\paragraph{Observed residual.}
At the first refinement iteration, the generated trajectories showed a substantial mismatch in daily stop counts, with a stop-count MAE of 4.663 and KL divergence of 6.656. The overall validation objective was 5.346. Although the simulator was executable, these discrepancies indicated that its daily activity generation mechanism was inconsistent with the observed mobility data.

\paragraph{Mechanism diagnosis.}
The Feedback Generation Agent (FGA) traced the residual to an inconsistency between the planned stop-count mechanism and end-of-day handling. The simulator first sampled a fixed number of activities from the learned stop-count distribution, but terminated generation when accumulated travel time pushed an arrival beyond the daily boundary. As a result, the sampled stop count could exceed the number of activities realizable within the remaining time budget. FGA identified this ``sample-then-truncate'' behavior as a source of systematic censoring.

\paragraph{Localized repair.}
Guided by this diagnosis, the subsequent Code Generation Agent (CGA) revision modified the simulator's daily activity generation mechanism rather than merely adjusting a numerical parameter. Before generating activities, the revised simulator estimated a feasible activity cap from the remaining daily time and expected travel-plus-dwell time per activity, and constrained the sampled stop count accordingly. It also aligned activity start times with travel feasibility by taking the later of the target activity time and simulated arrival time.

\paragraph{Result after repair.}
On the same N$\rightarrow$N validation set of 69 agents and 1,071 agent-days, the next iteration reduced stop-count MAE from 4.663 to 1.749 and KL divergence from 6.656 to 1.223. The overall objective decreased from 5.346 to 2.165. Time-of-day JSD decreased from 0.0912 to 0.0387, distance Wasserstein distance from 0.5918 to 0.3778, and transition divergence from 0.01068 to 0.00969, while top-$k$ POI recall increased from 0.2188 to 0.2855.

This example illustrates a diagnosis-consistent repair trace in which a behavioral discrepancy is localized to a simulator mechanism, translated into a targeted structural modification, and followed by improved behavioral fidelity. We treat it as a qualitative illustration rather than a single-edit causal ablation, since other refinements may also occur between successive iterations.